# Overtwisting and Coiling Highly Enhances Strain Generation of Twisted String Actuators

Revanth Konda[*1], David Bombara[*1], and Jun Zhang [*]

**Abstract**

Twisted string actuators (TSAs) have exhibited great promise in robotic applications by generating high translational force with low input torque. To further facilitate their robotic applications, it is strongly desirable but challenging to enhance their consistent strain generation while maintaining compliance. Existing studies predominantly considered overtwisting and coiling after the regular twisting stage to be undesirable—non-uniform and unpredictable knots, entanglements, and coils formed to create an unstable and failure-prone structure. Overtwisting would work well for TSAs when uniform coils can be consistently formed. In this study, we realize uniform and consistent coil formation in overtwisted TSAs, which greatly increases their strain. Furthermore, we investigate methods for enabling uniform coil formation upon overtwisting the strings in a TSA and present a procedure to systematically "train" the strings. To the authors' best knowledge, this is the first study to experimentally investigate overtwisting for TSAs with different stiffnesses and realize consistent uniform coil formation. Ultra-high molecular-weight polyethylene (UHMWPE) strings form the stiff TSAs whereas compliant TSAs are realized with stretchable and conductive supercoiled polymer (SCP) strings. The strain, force, velocity, and torque of each overtwisted TSA was studied. Overtwisting and coiling resulted in approximately 70% strain in stiff TSAs and approximately 60% strain in compliant TSAs. This is more than twice the strain achieved through regular twisting. Lastly, the overtwisted TSA was successfully demonstrated in a robotic bicep.

1. **Introduction**

The twisted string actuator (TSA) is an artificial muscle consisting of two or more strings connected to an electric motor on one end and a load on the other end.[1–3] As shown in Figs. 1a–b, actuation is realized by twisting the strings with a motor to linearly displace a load.[1–3] TSAs typically generate strains of approximately 30–35% of their untwisted length with high energy-efficiency.[2,3] Most commonly used soft actuators and artificial muscles exhibit one or more evident limitations such as (1) fabrication complexity,[4,5] (2) high power requirement,[6,7] (3) slow actuation,[8,9] and (4) insufficient force generation.[9–11] For example, dielectric elastomer actuators (DEAs)

This work was supported in part by a NASA Space Technology Graduate Research Opportunities award.
[*] Revanth Konda, David Bombara, and Jun Zhang are with the Department of Mechanical Engineering, University of Nevada, Reno, NV 89557, USA (rkonda@nevada.unr.edu; dbombara@nevada.unr.edu; jun@unr.edu)
[1] These authors contributed equally to this work and are considered to be co-first authors.

and hydraulically amplified self-healing electrostatic (HASEL) actuators, while generating sufficient actuation, require complicated and high-cost fabrication procedures that could be potentially challenging to realize.[5,7] DEAs and HASEL actuators also require high-voltage equipment (producing over 10 kV). This is not only costly but also potentially dangerous. Similarly, although pneumatic actuators exhibit appreciable strain and force generation, adopting them in robotic devices requires pumps or compressors.[12,13] Furthermore, the thermal actuation of shape-memory alloys (SMAs) not only results in low bandwidth and low energy efficiency, but also could be dangerous and cause damage to the robotic devices in which they are employed.[8,14] TSAs could present unique advantages over other actuation mechanisms. These advantages are in part due to TSAs' simplicity in fabrication and assembly, low power requirement, high operating frequency, large linear strain generation, and compactness.[1] Furthermore, by using stretchable and conductive strings, compliance and self-sensing properties could also be obtained.[15,16] TSAs have been used in tensegrity robots, robotic gloves, robotic hands, robotic exoskeletons, and other applications.[1]

The important performance metrics of a TSA are the contraction range, contraction speed, force output, torque input, bandwidth, and compliance. By utilizing a suitable pair of motor and strings, a TSA can be constructed to satisfy a given force output, bandwidth, and compliance requirement.[1] However, one area that is independent of the components of a TSA and needs further improvement is the strain generation (relative to the initial string length). Compliance is essential in areas where safe interaction with human beings is demanded. Previous work used elastic strings in a TSA-driven robotic joint,[17] whereas other work employed elastic supercoiled polymer (SCP) strings in TSAs.[15,16] The inclusion of SCP strings in TSAs has the major advantage of resistance based self-sensing behavior.[15,16] SCP strings have attracted much attention because they are compliant, lightweight, elastic, and often electrically conductive.[18–21] The electrical resistance in SCP strings has been found to vary according to the strings' elastic deformation.[22] In sensing of the TSA, it is desirable to estimate its strain based on an electrical signal of the actuator. Unlike many other soft actuators, TSAs are unique in that the actuator only consists of two or more strings connected to a motor with no constraints on the type or material of strings. Therefore, replacing the typically adopted strings with conductive SCP strings could enable self-sensing TSAs. However, the inclusion of SCP strings resulted in less strain compared to stiff strings: stiff string-based TSAs generate strain of up to 30–35% strain of their initial length, whereas SCP-based TSAs produce strains of around 20% of their initial length.[15,16] Although compliance and self-sensing are desirable, large strain generation is equally critical. Heating the SCP strings could result in additional strain of approximately 15% after they are fully twisted (making the total output strain around 35%),[15] but self-sensing while heating the strings has not been sufficiently studied because the electrical resistance is likely coupled with temperature, making the self-sensing behavior significantly difficult to characterize and model. Due to these reasons, the need for an alternative actuation mechanism which increases the strain output of the SCP-based TSAs, while not affecting their self-sensing capabilities, is strongly desirable.

In past studies, larger strains were obtained by "overtwisting" the strings—twisting beyond the regular twisting stage (Fig. 1c). In a study by Tavakoli et al.,[23] with a two-string configuration, up to 72%-strains were obtained by overtwisting strings at loads between 2-kg to 5-kg.[23] Although Tavakoli et al. claimed overtwisting to be advantageous, existing studies predominantly considered overtwisting to be undesirable.[24–28] This could be due to (1) increased nonlinearity and (2) the formation of non-uniform and unpredictable knots, entanglements, and coils in the overtwisted strings.[24,27,28] In the study by Tavakoli et al., although large strains were generated using overtwisting, the formation of coils was considered to be undesirable: they were neither uniform nor consistent.[6,8,17,18] Whereas the increased nonlinearity could be handled using advanced modeling techniques, the unpredictable non-uniform coil formation is most detrimental and undesirable. Firstly, the non-uniform coil formation accelerates the actuator's failure and causes unpredictable behaviors. Secondly, non-uniform coils occupy more space in the direction perpendicular to the actuation direction relative to uniform coils. This could result in unwanted interaction of the strings with other system components.[23] In the meantime, TSAs in general have attracted much attention in the robotics community, resulting in increased usage to power multiple advanced robotic and mechatronic devices.[27–30] It is thus strongly desirable to develop novel TSAs with increased strains.

Overtwisting could be an efficient mechanism for large strain generation for TSAs when uniform coils are consistently obtained. However, no detailed studies have been conducted to realize uniform coils in overtwisted strings with high repeatability. In our work we present a novel actuation mechanism which enables TSAs to generate very high strains predictably and repeatably. By applying the proposed actuation mechanism, uniform coil formation was obtained. Furthermore, we present a procedure to systematically "train" the strings to ensure they achieve uniform coil formation. Our experiments confirmed that both stiff and compliant TSAs can achieve uniform coils at a wide range of loading conditions. The main contributions of this paper are:

- Proposal of overtwisting and coiling in TSAs to greatly increase their consistent strain. The proposed method is experimentally validated on TSAs with different diameters, stiffnesses, and applied loads.
- Investigation of methods to enable TSAs to achieve repeatable uniform coil formation at large range of loading conditions.
- Comprehensive experimental characterization of the behavior of the proposed mode of actuation, and its demonstration in a standard robotic device.

To our best knowledge, this is the *first* study to experimentally investigate overtwisting for TSAs with different stiffnesses and realize consistent uniform coil formation. For stiff TSAs, ultra-high molecular-weight polyethylene (UHMWPE) strings with different diameters were studied. Compliant TSAs refer to those fabricated from SCP strings and thus will be used interchangeably throughout the paper. Through overtwisting and coiling, strains of up to 72% were obtained in stiff TSAs whereas strains of up to 69% were achieved in compliant TSAs.

2. **Materials and Methods**

2.1. *Proposed Mode of Actuation*

The proposed mode of actuation consists of two phases of twisting: the regular twisting phase and overtwisting phase. In the regular twisting phase,[2,3] the two strings were twisted together such that they formed a tightly packed double helical structure.[31] In the overtwisting phase, the tightly packed double helical structure then twisted to coil like single (thicker) string. Uniform coils formed along its length, resulting in the formation of a tightly packed helical structure.[31] It is noted that the proposed method is based on uniform coil formation upon overtwisting. The transition of the strings from a fully twisted state to the overtwisting state and an example of non-uniform coils are presented in Fig. 2.

2.2. *Training of Stiff TSAs*

Experiments demonstrated that TSAs with unused SCP strings consistently produced uniform coils during overtwisting, but TSAs with unused stiff strings did not. Initially it was found that uniform coil formation in TSAs with stiff strings could be obtained only at sufficiently high loading conditions. This observation was consistent with previous studies on coil formation in artificial muscles.[32,33] However, uniform coil formation at lower loads—and thus a wider loading range—is desirable. Upon further investigation, the strings were found to exhibit repeatable and uniform coil formation through sufficient "training" at low loads. Although this type of training process is common for other smart materials, this procedure has never been reported for TSAs.

The training procedure consisted of the following steps: Firstly, based on the strings' diameter, a physically feasible minimum load that ensured the strings to be sufficiently taut was chosen. "Sufficiently taut" is a condition where, after fully twisted in the normal twisting phase, the strings can return to their initial state upon untwisting without requiring any additional external force besides the applied payload. In this study, it is assumed that if a particular load ensured that the strings were sufficiently taut during the normal twisting phase, then the load would also ensure sufficient tautness for the overtwisting phase. Secondly, at this load, the strings underwent twisting and untwisting cycles until uniform coils formed. 50 cycles were often sufficient to train the TSA with stiff strings to generate uniform coils during overtwisting. As an example, the evolution of coil structure is shown in Fig. 3, for the TSA with 1-mm UHMWPE strings under 200-g load. This evolution is described as follows: In the first stage, the coiling took place in the direction perpendicular to the linear actuation (Fig. 3a). In the second stage, some coils were formed along the length of the strings and remaining coils projected in the perpendicular direction (Fig. 3b). In the third stage, the coil formation started taking place along the length of the strings, however, the coils' diameters were uneven (Figs. 3c-d). In the final stage, uniform coils formed inline with the strings of the TSA (Fig. 3e). This procedure caused mild softening and gradual deformation of the strings into a helical shape. Although this deformation caused a slight decrease of the TSA's operating length, it enabled the strings in each TSA to consistently form uniform coils. Since the strings in TSAs were trained at the physically feasible minimum load, they were able to form uniform coils at higher loads as well. The training process is critical for stiff strings as it enabled them to achieve uniform coil formation upon overtwisting and consequently enabled stiff TSAs to attain high strains at wide ranges of loads. More details on the training process can be found in the supplementary material S1.4.

### 2.3. Uniformity of Coils

In this paper, for stiff strings, uniformity of coils implied homogeneity of the coils. Homogeneity of coils suggests that all the coils have approximately the same size and shape. Although regularity of the coils is also an important criterion, it was relaxed because during training of stiff strings, it was observed that while most coils were regular, there were few coils which formed at slightly irregular intervals (Fig. 3e). TSAs with compliant SCP strings did not require training to achieve uniform coil formation. This is likely due to their low stiffness and their supercoiled structure. However, even in SCP strings, while homogeneity of coils was observed, some coils formed with irregular spacing between them, similar to stiff strings.

### 2.4. Performance Metrics

In this work, the strain $S$ of the TSA is defined to be negative when the TSA's length decreased, whereas the contraction $C$ is the negative of the strain. The strain is expressed as

$$S = -C = \frac{L - L_0}{L_0} \times 100\%,$$

where $L$ is the current length of the TSA and $L_0$ is the initial loaded length. For stiff TSAs, $L_0$ corresponded to zero motor rotations. Since the experiments with compliant TSAs began with the TSA under 10 rotations, $L_0$ corresponded to 10 rotations.

3. **Results**

*3.1. Stiff TSAs*

The TSA with 1.0-mm, 1.3-mm, and 2.0-mm UHMWPE strings, which were loaded with 2000-g, 2900-g, and 3400-g, respectively, were used to experimentally study multiple properties of the overtwisting actuation mode. Beyond these loads, the TSAs were likely to fail, or the motor was likely to stall.

Firstly, the strain generation was characterized for the stiff TSAs. The input sequence applied to the TSA is presented in Fig. 4a and the corresponding strain of the stiff TSA with 1.3-mm-diameter strings is presented in Figs. 4b–c. The motor angle–strain relationship was mildly hysteretic, but the hysteresis was negligible in the regular twisting phase. This could be because of the strings' high stiffness and load. Similar trends were found for other stiff TSAs. In all cases, around 70% contraction was observed consistently (Figs. 4b–c and Fig. S4).

Secondly, the actuation velocity of stiff TSAs was obtained. The actuation velocity of the stiff TSA with 1.3-mm strings was calculated from the strain–motor rotations correlation in Fig. 4b and the velocity is presented in Fig. 4d. For all the stiff TSAs, the actuation speed was significantly greater in the overtwisting phase than in the regular twisting phase. This implies that the bandwidth of the TSA's strain is greater during overtwisting than during normal twisting. More details on the bandwidth can be found in the supplementary material S2.4.

Thirdly, the motor torque of the stiff TSAs was studied. This torque was computed by multiplying the current of the DC motor by the motor torque constant. The results of stiff TSA with 1.3-mm-diameter strings are presented in Fig. 4e. The results indicated that overtwisting in stiff TSAs required more torque than the regular twisting phase.

A summary of the maximum contractions, actuation speeds, and required torques of different stiff TSAs are presented in Figs. 4f, 4g, and 4h, respectively. Numerical results are also presented in Table 1. Full results are found in Figs. S5–S7 and Supplementary Video S1. Results are also presented for TSAs made from Kevlar strings in Supplementary Fig. S8 and Supplementary Video S2.

Finally, a test of the TSA's lifetime was conducted (Fig. 4i). A TSA with 1.3-mm-diameter UHMWPE strings under 2900-g of load was used. The motor rotated with five-rotation steps and a 200-ms pause between each rotation. The TSA lasted for 1030 cycles without breaking. The TSA consistently achieved approximately 60% contraction. Although the overall length of the TSA slightly decreased over time, the linear actuation range remained approximately constant between all cycles. The strings also gradually deformed into a helical shape. Consequently, the length of the TSA decreased, as verified by the creep-like behavior in Fig. 4i. As the experiment ran, the number of coils that formed along the length of the strings also increased. Furthermore, while the generated strain changed initially due to the decrease in length of the strings, after approximately 2 hours from the beginning of the test, the strain did not further increase and was consistent.

*3.2. Compliant TSAs*

As an example, the strain and self-sensing of 6-ply SCP based-TSA under a 200-g load is considered. As shown in Figs. 5a–b, the TSA repeatably achieved 11.25% strain in the regular twisting phase, then an additional 46.89% in the overtwisting phase. The relationship between twists and strain distinctly revealed the regular twisting and overtwisting phases of the TSA, as shown in the two hysteresis loops in Fig. 5b. Compliant TSAs demonstrated greater hysteresis than the stiff TSAs in both the normal twisting and overtwisting phases, mainly due to the friction and material properties of SCP strings. Because the TSA was fabricated from conductive SCP strings, the electrical resistance of the strings was measured. Transient decays in the electrical resistance were observed (Fig. 5c), similar to transient resistance behaviors in other materials.[34,35] The results suggest a correlation between the strain and resistance of the TSAs, which could be used for strain self-sensing. The resistance experienced mild creep as time increased, which was likely due to the friction between the strings that rubbed the silver coating off of them.[16]

A summary of the maximum repeatable contractions for the compliant SCP-based TSAs is shown in Fig. 5d. Full results are provided in Figs. S9–S10 and Supplementary Video S3. The strains achieved by SCP-based TSAs in regular twisting phase were less than those achieved by stiff TSAs mainly because the SCP strings are compliant and stretchable.[15] Increased loads generally decreased the maximum attainable contraction (Fig. 5d). Furthermore, as the load increased, the contraction obtained through regular twisting increased but the contraction obtained through overtwisting decreased.

The lifetime endurance was also examined in the compliant TSA. Two tests were conducted with different pairs of 6-ply SCPs loaded with 100-g. On the first test, the TSA lasted for 285 cycles with consistent 60% contractions but broke on the 286[th] cycle (Fig. 5e). On the second test, the TSA lasted for 428 cycles and achieved consistent contractions of 45% (Fig. 5f). For both tests, the resistance demonstrated creep over time. This could be due to the friction between the strings that rubbed the silver coating off them. The TSA in the second test achieved less contraction and less resistance change because the maximum twisting angle was smaller.

*3.3. Discussions*

The proposed mode of actuation significantly improves the strain generation of TSAs and increases their impact in the field of robotics, more specifically in applications demanding high force and compact actuators, such as robotic exoskeletons,[36,37] tendon-driven robotic fingers and arms,[28] and other robotic structures.[38,39] Large strains generated through overtwisting and coiling could result in the realization of compact solutions in areas that demand large displacements. For example, consider a robotic device like the one presented in Hosseini et. al.,[37] which requires displacements of 10-mm. To meet this requirement, if only regular twisting is employed (with up to 30% contraction), then a TSA with a length of approximately 33-mm would be required. However, with overtwisting and coiling (with up to 70% contraction), a TSA with a length of only 14.3-mm would be sufficient. This would effectively reduce the untwisted strings of the TSA to less than half of their original length. Even in the overtwisted phase, the twisted strings only mildly increased in diameter, which still allows the actuator to be conveniently placed in conduit or a sheath for wearable and assistive devices.[40,41] For example, a TSA made from two 6-ply SCP strings and loaded with 200-g hanging weight had a diameter of 2.1-mm in the regular twisting phase and only 3.6-mm in the overtwisting phase. The uniformly coiled strings can also be put into bending actuators,[42] and then can be further incorporated into different robotic applications. Furthermore, soft robotic structures, such as soft bending fingers,[10] could be realized using compliant, overtwisted, and uniformly coiled TSAs. Such devices would be able to generate larger forces and quicker actuation while maintaining compliance and compactness.

Some evident limitations of the proposed mode of actuation are as follows: Firstly, the uniform coil formation, while enabling a stable and repeatable behavior, resulted in jerky motion due to formation of coils. Although the coils were uniform, every time a new coil was formed, the string would intermittently contract at a faster rate. This may generate unwanted oscillations of the payload attached to the string. Secondly, the proposed overtwisting mechanism increased the nonlinearity of the actuator. Hysteresis in TSAs and particularly compliant TSAs made with SCP strings has been recently studied in our previous work.[15,16] Our previous study on self-sensing SCP-based TSAs showed that the length-resistance correlation is not only hysteretic but also contains creep and transient decay.[16] While the hysteresis in compliant TSAs is significant, stiff strings have relatively less hysteresis. This may be due to the high stiffness of the strings.

Furthermore, previous studies indicate that even in the regular twisting phase, TSAs exhibit hysteresis.[3] In addition it was observed that the hysteresis in the overtwisting phase became less significant with increasing payload. This was observed in both stiff and compliant TSAs. Overall, few studies have been conducted to study the hysteresis properties of TSAs, including the characterization, modeling, and control.

Hysteresis is an extensively studied nonlinearity in the robotics and mechatronics community. To capture hysteresis, both physics-based methods and phenomenology-based approaches, such as the Preisach operator, Prandtl-Ishlinskii model, and Bouc-Wen model, have been proposed.[11,43–46] Furthermore, different control schemes like feedforward control, proportional–integral–derivative control, robust control, and adaptive control have been realized.[47–50] In addition, the actuator also exhibited creep-like behavior as can be inferred from Fig. 5i. Previous studies on TSAs with regular twisting have hinted creep-like behavior which was observed in the current study as well.[51] Creep is a common nonlinearity found in many artificial muscles and smart materials, which does not make the actuator unstable if treated appropriately. Like hysteresis, creep has been extensively modeled in previous work.[52,53] Those models could apply to overtwisting as well.

Lastly, compliant TSAs exhibited low lifetimes in comparison to stiff TSAs. This is mainly due to the inherent limitations of SCP strings: The coiled structure of the SCP strings resulted in increased frictional forces between the strings when twisted and coiled. Consequently, the wear and tear of the strings increased which led to their early failure. Meanwhile, the self-sensing realized by using conductive strings is an important concept because of the compact sensing achieved through it. The lifetime of complaint self-sensing TSAs can be potentially enhanced by using robust stretchable conductive strings.

### 3.4. Robotic Demonstration

Overtwisting and coiling of TSAs has great potential in robotic exoskeletons,[36,37] assistive devices,[40,41,54] tendon-driven robotic fingers and manipulators,[28,55–57] and other robotic structures[38,39,58]. In this paper, a TSA-driven robotic bicep demonstrated the effectiveness of overtwisted TSAs. 1.3-mm UHMWPE strings were used. The device was selected to demonstrate the effectiveness of the proposed mode of actuation because it has been employed by other similar works[5,11] to demonstrate the capabilities of novel soft actuators and artificial muscles. Furthermore, the robotic bicep is different from the weight-lifting setup used for experimental characterization because the tension force experienced by the strings changes as the bicep angle changes. This application therefore shows the enhanced strain and force outputs of the overtwisted TSAs simultaneously. The robot was constructed using three metallic rods. The first rod (the stationary link, labeled as the upper arm in Fig. 6a) was suspended vertically from a metallic frame. The other two rods were assembled parallel to each other and formed the movable link, labeled as the forearm in Fig. 8a. One end of the forearm was attached to the bottom of the upper arm by a revolute joint. The free end of the forearm was loaded with 700-g. A potentiometer at the joint measured the bending angle relative to the initial position of the forearm (Fig. 6a). The TSA was installed with the motor attached to the upper arm and the bottom of the strings attached to the forearm. The initial angle was $\phi_1 = 13.1°$. Whereas regular twisting of the TSA resulted in a maximum bending angle of $\phi_2 = 73.4°$ ($\phi_2 - \phi_1 = 60.3$, Fig. 6b), overtwisting resulted in the maximum bending angle of $\phi_3 = 147.10°$ ($\phi_3 - \phi_2 = 73.7°$, Fig. 6c). The bending angle versus motor rotations is presented in Fig. 6d. The untwisted, fully twisted, and overtwisted lengths of the TSA were 21.5-cm, 13.5-cm, and 6.8-cm, respectively. Supplementary Video S4 depicts the bicep in motion.

4. **Conclusion and Future Work**

In this study, the overtwisting and coiling of stiff and compliant TSAs were experimentally investigated. In both cases, large strains of more than twice the strains achieved by regular twisting were obtained. The findings were further supported by the actuation speed, force generation and required torque of the TSAs. The consistent electrical resistance measurements indicated that the overtwisted compliant TSAs have strong potential to be utilized for strain self-sensing.

Future work will overcome the limitations of the overtwisting phase. The first limitation is that the number of twists must be precisely monitored to achieve maximum strain without breaking the TSA. The second limitation is that the friction experienced by the coils during overtwisting wears the silver coating off from the nylon threads at a higher rate (than during normal twisting), thereby affecting their self-sensing behavior. In future work, firstly, a physics-based model of overtwisting and coiling of the TSA will be developed. Consequently, closed-loop position control will be realized. Secondly, high-endurance self-sensing could be realized in stiff and compliant TSAs by coating the strings with high-performance conductive material, similar to work by Piao et al.[60]

**Acknowledgements**

The authors would like to thank Cianan Brennan at the University of Nevada, Reno for his assistance in fabricating the robotic bicep.

**Author Contributions**

J.Z. conceived the idea. R.K. and D.B conducted experiments and analysis. All authors wrote the manuscript.

**Author Disclosure Statement**

No competing financial interests exist.

**Supplementary Material**

Supplementary Text S1

Supplementary Text S2

Supplementary Figure S1

Supplementary Figure S2

Supplementary Figure S3

Supplementary Figure S4

Supplementary Figure S5

Supplementary Figure S6

Supplementary Figure S7

Supplementary Figure S8

Supplementary Figure S9

Supplementary Figure S10

Supplementary Video S1

Supplementary Video S2

Supplementary Video S3

Supplementary Video S4

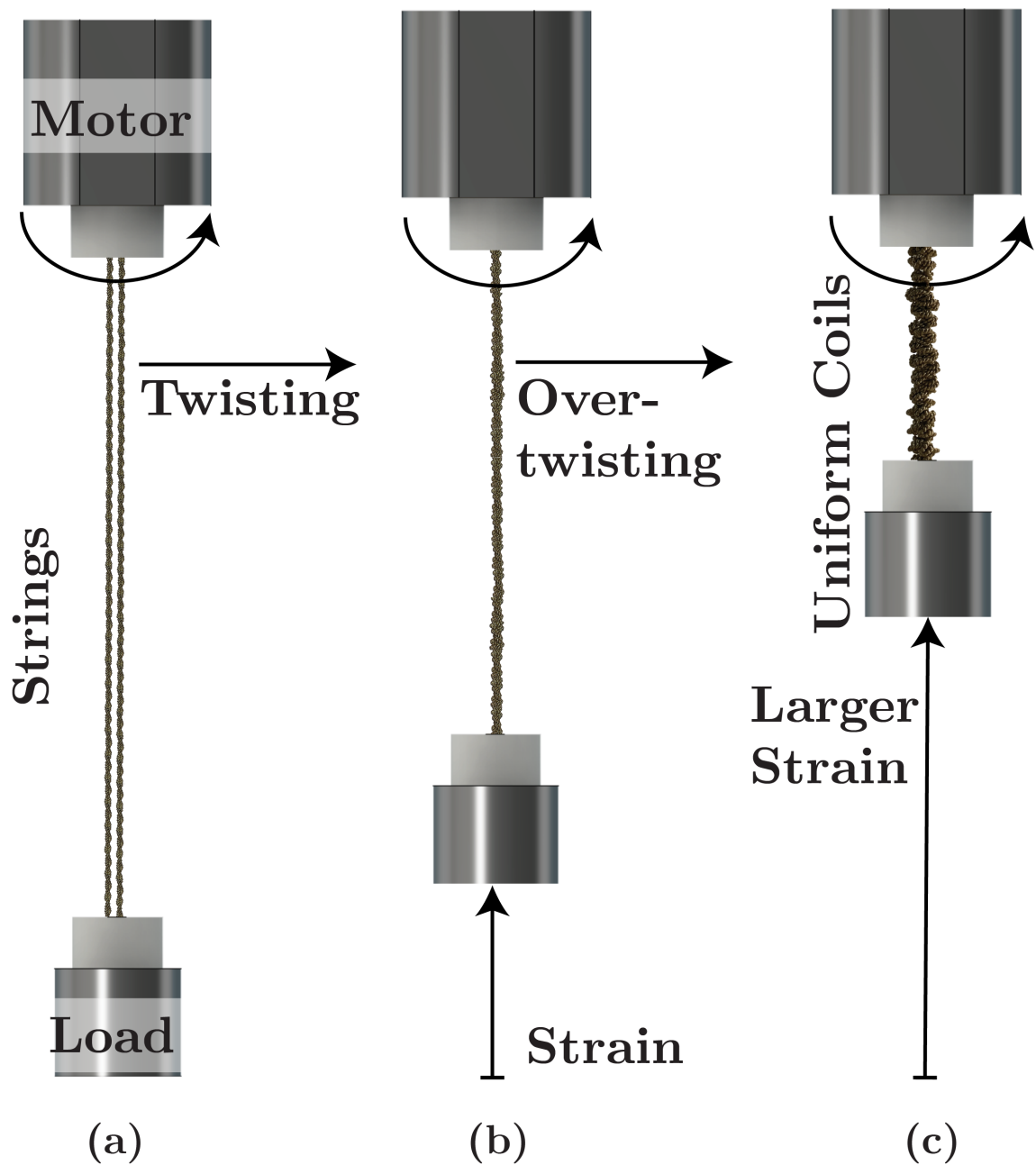

*Figure 1: The (a) untwisted twisted string actuator (TSA), (b) TSA in the regular twisting phase, and (c) completely overtwisted and coiled phase.*

18  KONDA AND BOMBARA *ET AL*

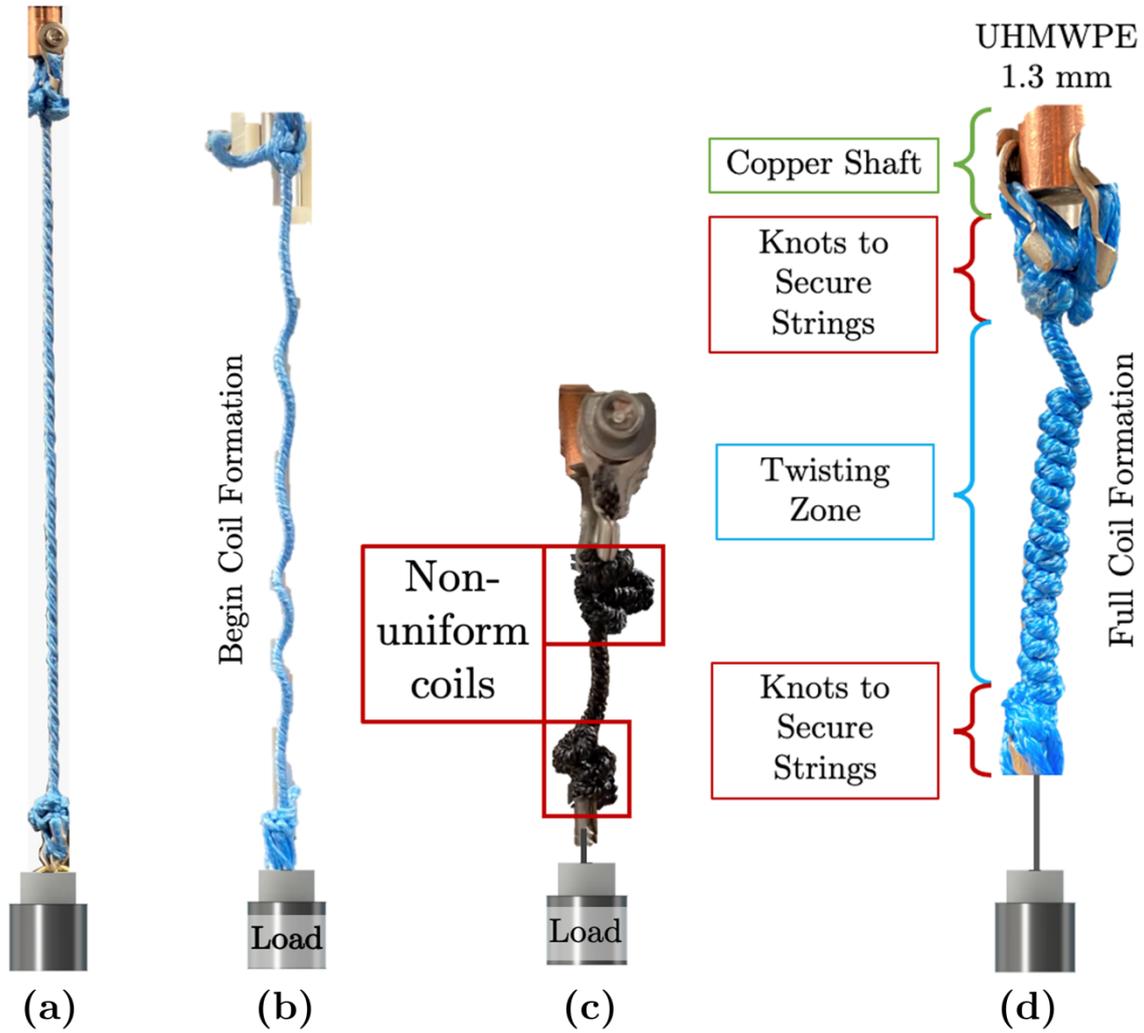

*Figure 2: (a) The regular twisting phase. (b) The transition of the strings from the fully twisted state to the overtwisted state. (c) The non-uniform knots, entanglements, and coils that were formed in stiff TSAs when the TSA was not pre-trained. (d) The uniform coils that were obtained by overtwisting trained TSAs.*

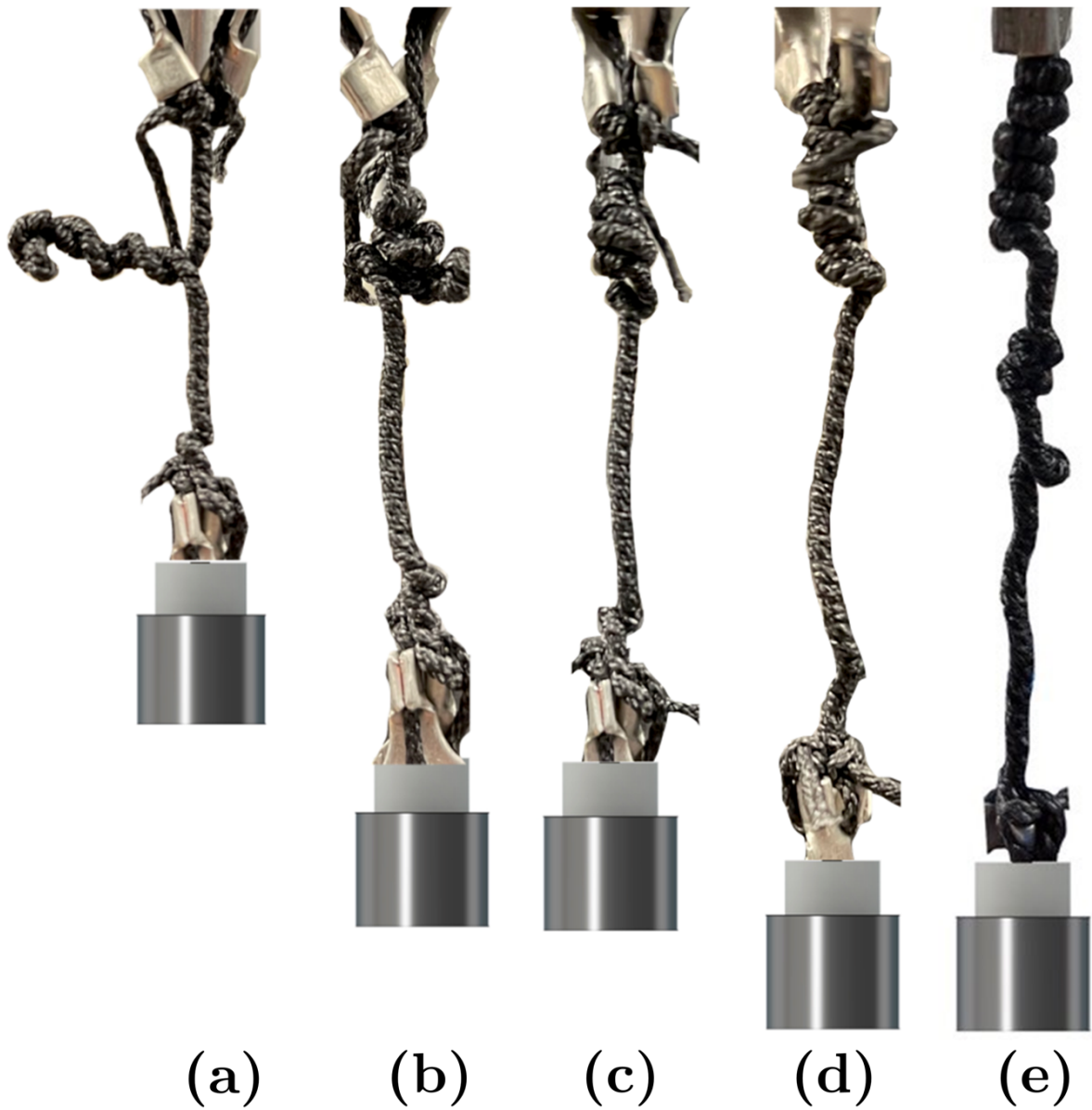

*Figure 3: The evolution of the coiling structure of 1-mm UHMWPE strings during training. The coiling structure of the TSA during the (a) 1st, (b) 6th, (c) 11th, (d) 18th, and (e) 50th cycle.*

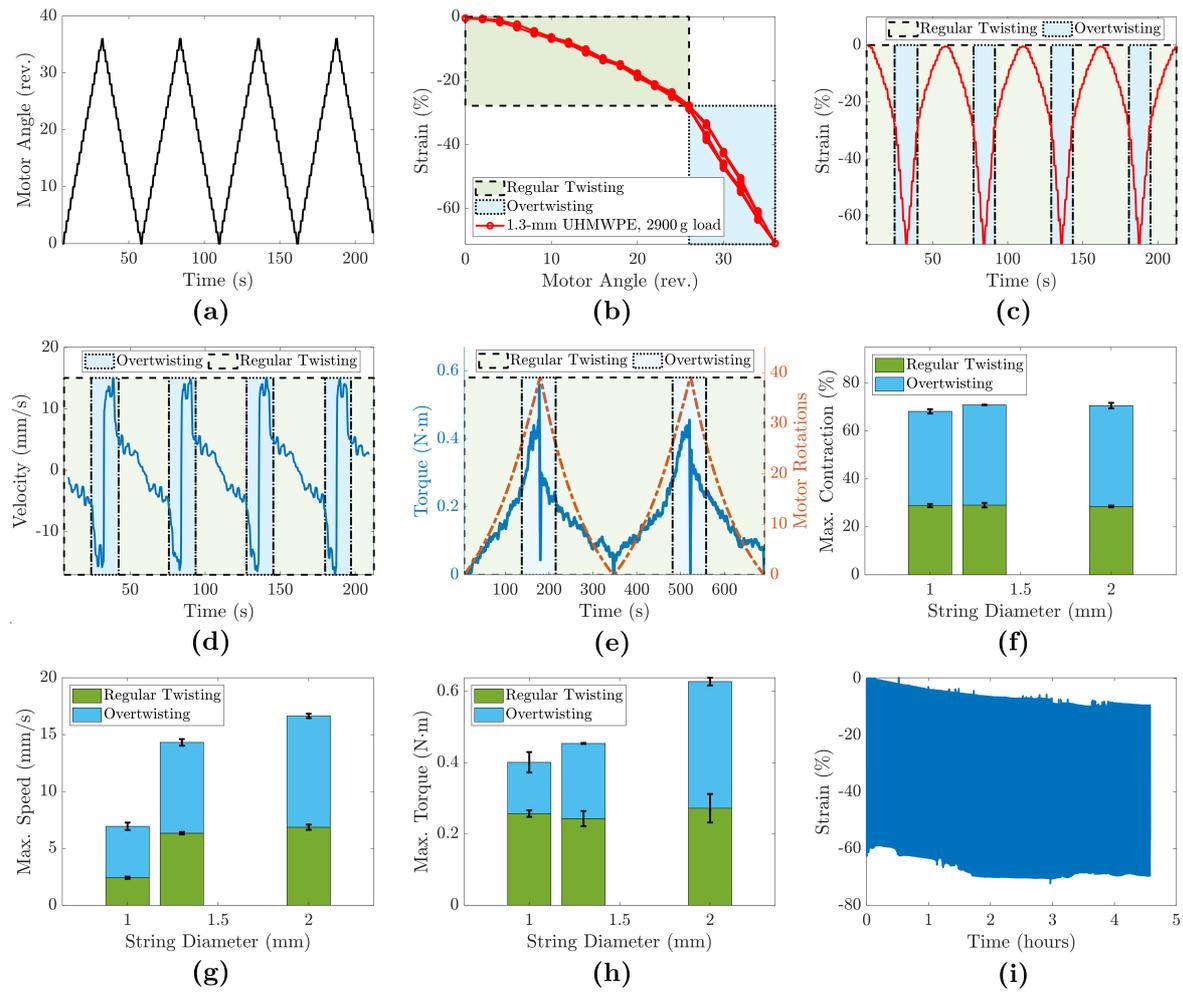

*Figure 4: Experimental results for the stiff TSAs. (a)—(e): Results obtained with 1.3-mm UHMWPE strings loaded with 2900g. The areas of the regular-twisting phase and the overtwisting phase are highlighted accordingly. (a) The motor angle input sequence versus time. (b) The correlation between strain and motor angle. The (c) strain, (d) velocity, and (e) input torque of the TSA versus time. Note that in (e), a motor angle input sequence with low angular velocity was utilized to decrease the noise in the torque measurement. (f)—(h): The comparison of results for UHMWPE strings with different diameters. Error bars are shown in black in each plot. The error bars represent the standard deviation of the maximum attained values over four cycles. The heights of the bars denote the maximum values that were averaged over four cycles. The (f) maximum contraction, (g) maximum speed, and (h) maximum input torque of the TSAs in the regular twisting and overtwisting phases. (i) A test of the TSA's lifetime by measuring the strain versus time for 1.3-mm UHMWPE strings.*

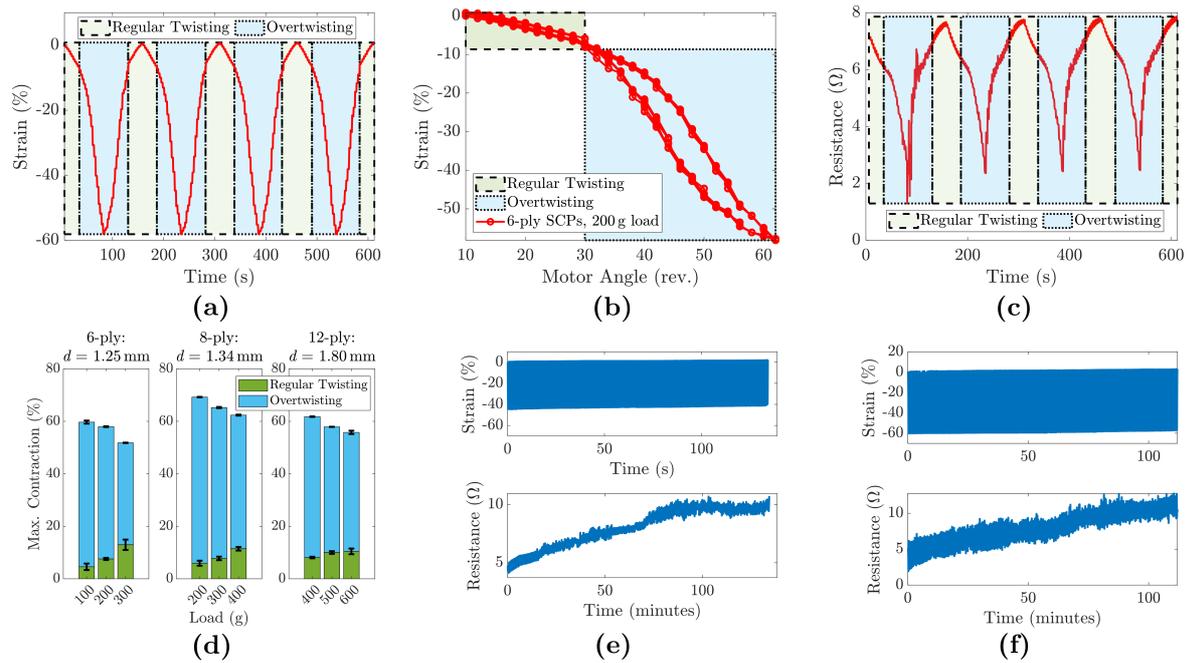

*Figure 5: The experimental results for the compliant SCP-based TSAs. The data presented in (a)–(c) was for 6-ply SCP-based TSA with a 200-g load. (a) The variation of strain in time. (b) The length-motor angle (turns) correlation. (c) The variation of resistance in time. (d) The maximum repeatable contractions obtained in compliant SCP-based TSAs. (e)–(f) The strain and electrical resistance of the overtwisted compliant SCP-based TSA over time. (e) For 428 complete cycles, the maximum contraction consistently reached approximately 45%. (f) For 285 cycles, the TSA achieved consistent strains of approximately 60%.*

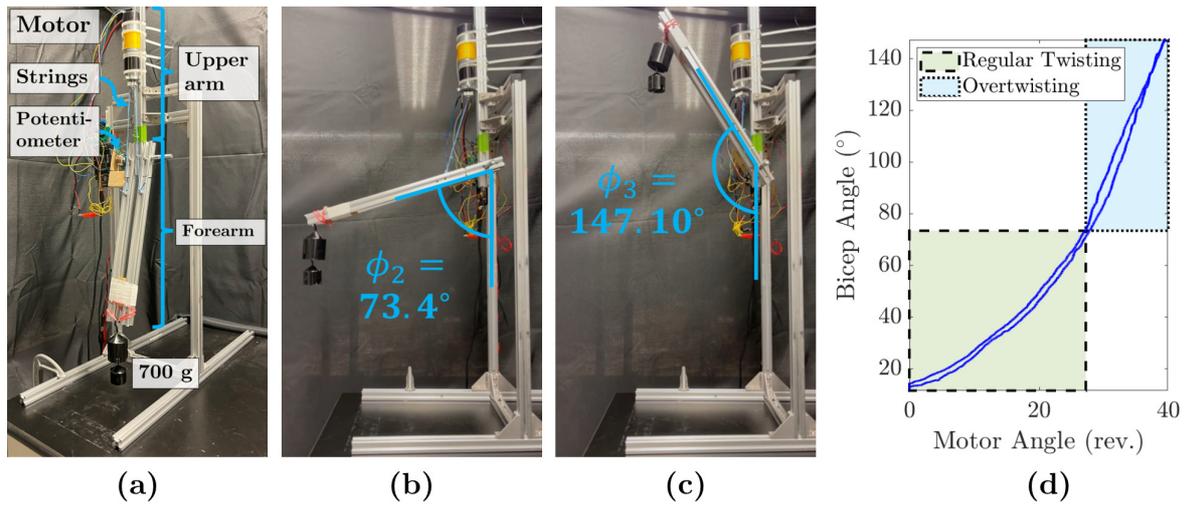

*Figure 6: (a) Robotic bicep actuated by TSA with UHMWPE strings having diameter of 1.3-mm. (b) Position of the forearm at the end of normal twisting phase. (c) Position of the forearm achieved through overtwisting and coiling. (d) The bending angle versus motor rotations for the robotic bicep.*

Table 1: Comparison of Overtwisted TSAs with stiff strings.

| Diameter (mm) | | 1.0 | 1.3 | 2.0 |
|---|---|---|---|---|
| Initial Length (cm) | | 22.42 | 21.43 | 25.32 |
| Maximum Load (g) | | 2000 | 2900 | 3400 |
| Maximum Motor Angle (rev.) | | 56 | 36 | 24 |
| Maximum Contraction (%) | Regular Twisting | 28.90 | 29.08 | 28.53 |
| | Overtwisting and Coiling | 68.22 | 70.94 | 70.63 |
| Maximum Actuation Speed (mm/s) | Regular Twisting | 2.42 | 6.34 | 6.86 |
| | Overtwisting and Coiling | 6.94 | 14.32 | 16.65 |
| Maximum Required Torque (N·m) | Regular Twisting | 0.257 | 0.243 | 0.272 |
| | Overtwisting and Coiling | 0.401 | 0.454 | 0.627 |